\documentclass[runningheads]{llncs}

\usepackage{eccv}

\usepackage{eccvabbrv}

\usepackage{graphicx}
\usepackage{booktabs}
\usepackage{tabularx}

\usepackage[accsupp]{axessibility}  % Improves PDF readability for those with disabilities.

\usepackage{hyperref}

\usepackage{orcidlink}

\begin{document}

% ---------------------------------------------------------------
% TODO REVIEW: Replace with your title
\title{Competitive Memory Readout for Robust Video Object Segmentation: 2nd Place Technical Report for the MOSEv2 Track of the 8th LSVOS Challenge}

% TODO REVIEW: If the paper title is too long for the running head, you can set
% an abbreviated paper title here. If not, comment out.
\titlerunning{2nd Place Technical Report for the MOSEv2 Track}

% TODO FINAL: Replace with your author list. 
% Include the authors' OCRID for the camera-ready version, if at all possible.
\author{Mingqi Gao\inst{1} \and
Sijie Li\inst{1} \and
Jungong Han\inst{2}}

% TODO FINAL: Replace with an abbreviated list of authors.
\authorrunning{M. Gao et al.}
% First names are abbreviated in the running head.
% If there are more than two authors, 'et al.' is used.

% TODO FINAL: Replace with your institution list.
\institute{School of Computer Science, University of Sheffield\\
\email{\{m.gao,sli256\}@sheffield.ac.uk} \and
Department of Automation, Tsinghua University\\
\email{jghan@tsinghua.edu.cn}}

\maketitle
\setcounter{footnote}{0}

\begin{abstract}
We present our solution for the MOSEv2 track of the 8th Large-scale Video Object Segmentation (LSVOS) Challenge at ECCV 2026. The challenge evaluates robust video object segmentation under complex temporal dynamics, including long-term occlusion, disappearance and reappearance, large appearance changes, and strong interference from visually similar objects. Our method builds on SAM~3 and focuses on its memory readout. Standard target-only memory retrieval can confuse the annotated target with same-class non-target objects because such distractors are represented only implicitly as background. Our method introduces Competitive Memory Readout, which explicitly incorporates same-class competitor evidence when retrieving target information from memory. To prevent excessive suppression of weak or reappearing targets, we further apply a lightweight adaptive restoration rule after competition. The resulting system retains the original SAM~3 tracking pipeline while improving target identity preservation in challenging videos. Our submission achieves 66.20 on the primary challenge score and ranks 2nd in the MOSEv2 track.
\end{abstract}

\section{Introduction}
\label{sec:intro}

The 8th Large-scale Video Object Segmentation (LSVOS) Challenge\footnote{\url{https://lsvos.github.io/}}, held in conjunction with ECCV 2026, focuses on video segmentation under challenging real-world conditions. It contains three complementary tracks: MOSEv2, MeViSv2-Text, and MeViSv2-Audio. The MOSEv2 track studies complex semi-supervised video object segmentation, while MeViSv2-Text and MeViSv2-Audio consider referring motion expression video segmentation guided by textual and audio descriptions, respectively. Together, these tracks cover RGB-only, text-guided, and audio-guided video segmentation, providing a broad benchmark for object-centric video understanding under difficult temporal dynamics. Our submission focuses on MOSEv2~\cite{ding2025mosev2}, where the target mask is provided in the first frame and the model must segment the same object throughout the remaining video. Compared with earlier benchmarks such as DAVIS~\cite{perazzi2016benchmark} and YouTube-VOS~\cite{xu2018youtube}, the MOSE family places greater emphasis on realistic challenges including heavy occlusion, target disappearance and reappearance, crowded scenes, large appearance changes, and strong interference from visually similar objects~\cite{ding2023mose,ding2025mosev2}. These conditions make target identity preservation a central difficulty: even when the predicted masks remain visually plausible, the tracker can gradually drift from the annotated target to a same-class distractor.

Memory-based VOS has become the dominant paradigm for maintaining target information over time. Space-Time Memory Networks~\cite{oh2019stm} and XMem~\cite{cheng2022xmem} retrieve target-conditioned information from previous frames, while SAM~2~\cite{ravi2025sam2} integrates memory-based propagation into a general promptable video segmentation framework. SAM~3~\cite{carion2026sam3} further introduces concept-aware detection, segmentation, and tracking, providing a strong basis for challenging VOS. However, standard memory retrieval remains largely target-centric: the annotated target is explicitly represented in memory, whereas other regions, including same-class distractors, are treated only as non-target context. As a result, memory attention mainly measures whether a current region matches the target history, rather than whether it matches the target better than plausible alternatives. This limitation is particularly harmful in MOSEv2, where similar objects frequently coexist and can attract historical target memory after occlusion or reappearance. We therefore introduce \textbf{Competitive Memory Readout}, which explicitly compares target evidence against same-class competitor evidence during memory retrieval. A lightweight adaptive restoration rule is further used to recover valid target evidence when competition becomes overly suppressive. The resulting method retains the original SAM~3 tracking pipeline while directly targeting identity drift and weak target recovery, and achieves a primary challenge score of 66.20, ranking \textbf{2nd} in the MOSEv2 track.

\section{Related Work}
\label{sec:related}

Semi-supervised video object segmentation aims to propagate specified target objects through a video from first-frame masks~\cite{gao2023vosreview}. Early approaches relied on online adaptation or direct mask propagation, while memory-based methods became dominant by storing previous features and masks and retrieving them for current-frame prediction. STM~\cite{oh2019stm} established the space-time memory formulation, XMem~\cite{cheng2022xmem} improved long-term memory organisation, and Cutie~\cite{cheng2024cutie} introduced stronger object-level representations. More recently, SAM~\cite{kirillov2023sam} and SAM~2~\cite{ravi2025sam2} brought large-scale promptable segmentation to images and videos, leading to extensions such as SAM2Long~\cite{ding2025sam2long}, SAMURAI~\cite{yang2026samurai}, DAM4SAM~\cite{videnovic2025dam4sam}, and SeC~\cite{zhang2026sec}, which improve long-video memory handling, motion awareness, distractor robustness, or concept reasoning. SAM~3~\cite{carion2026sam3} further adds concept-aware detection and tracking. While these methods mainly improve memory selection, update, or maintenance, our method focuses on the readout itself by introducing same-class competitors as explicit evidence during target retrieval.

\section{Method}
\label{sec:method}

\subsection{Overview}

Our method builds on the standard SAM~3 video segmentation pipeline and keeps the backbone, memory encoder, and mask decoder unchanged. Given the first-frame target mask, SAM~3 propagates the object by retrieving target-conditioned information from stored memory. Our main modification is Competitive Memory Readout (CMR), which augments target-only retrieval with same-class non-target competitors. The key motivation is that a region can be highly similar to the target history while still corresponding to the wrong instance. CMR therefore calibrates target evidence against plausible alternatives rather than treating all non-target regions as undifferentiated background. Because competition can occasionally suppress a weak but correct target, especially after long occlusion or disappearance, we further apply a lightweight adaptive restoration rule. The two components are complementary: competition improves target discrimination, while restoration preserves recoverability.

\begin{figure}[t]
    \centering
    \includegraphics[width=.9\linewidth]{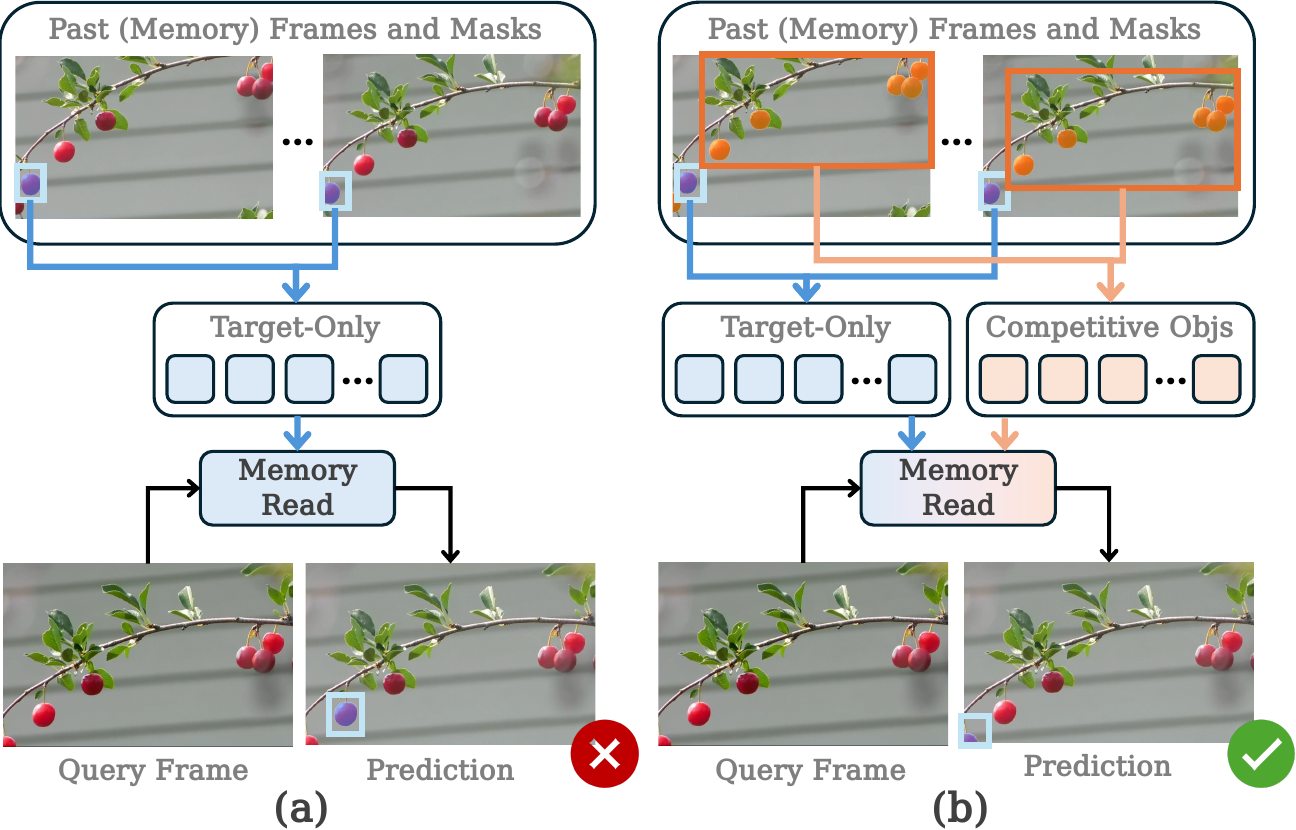}
    \caption{Overview of the competitive memory readout.
(a) Standard target-only memory readout retrieves target evidence without modelling visually similar non-target objects, which can lead to identity drift.
(b) Our method additionally introduces same-class non-target objects as competitors during memory readout, suppressing distractor-supported target responses and improving target identity preservation.
}
    \label{fig:overview}
\end{figure}

\subsection{Competitive Memory Readout}

For a target object $o$, let $\mathcal{T}^{o}_{f}$ denote its foreground memory tokens in memory frame $f$. Using the concept-aware detections provided by SAM~3, we collect same-class non-target hypotheses and remove those that strongly overlap the tracked target. The remaining regions are encoded as competitor foreground tokens $\mathcal{D}^{o}_{f}$. For each current-frame token $i$, we aggregate target and competitor attention logits using log-sum-exp,
\begin{equation}
r^{T}_{i,f}
=
\operatorname{LSE}_{u\in\mathcal{T}^{o}_{f}}\ell^{f}_{i,u},
\qquad
r^{C}_{i,f}
=
\operatorname{LSE}_{u\in\mathcal{D}^{o}_{f}}\ell^{f}_{i,u},
\end{equation}
and compute a competition gate
\begin{equation}
g_{i,f}
=
\sigma\left(
\frac{r^{T}_{i,f}-r^{C}_{i,f}}{\tau_c}
\right).
\end{equation}
The gate is applied to target foreground logits before the original memory-attention normalisation:
\begin{equation}
\widetilde{\ell}^{f}_{i,u}
=
\ell^{f}_{i,u}
+
\log(g_{i,f}+\epsilon),
\qquad
u\in\mathcal{T}^{o}_{f}.
\end{equation}
Thus, target evidence is preserved when a query region is better explained by the target memory and suppressed when a same-class competitor provides stronger evidence. If no valid competitor is available, CMR naturally reduces to the original target-only readout.

\subsection{Adaptive Restoration}

Competitive suppression improves identity preservation but can also reduce useful target evidence when the true target is weak or reappears after a long interval. Let $m^{f}_{\mathrm{pre}}$ and $m^{f}_{\mathrm{post}}$ denote the target response before and after competition for memory frame $f$. We measure the relative suppression strength and restore part of the target evidence using
\begin{equation}
s_f
=
\operatorname{clip}
\left(
\frac{m^{f}_{\mathrm{pre}}-m^{f}_{\mathrm{post}}}
{m^{f}_{\mathrm{pre}}+\epsilon},
0,1
\right),
\qquad
\rho_f = 1 + 0.5\sqrt{s_f}.
\end{equation}
The restoration factor lies between $1.0$ and $1.5$, providing little correction when competition is weak and stronger recovery when the target response is heavily suppressed. This deterministic restoration is applied before final attention normalisation and requires no additional learned policy.

\section{Experiments}
\label{sec:experiments}

\subsection{Challenge Setting}

We evaluate our method on the MOSEv2 track of the 8th LSVOS Challenge. MOSEv2 contains complex videos with frequent occlusion, disappearance and reappearance, large target variations, and multiple similar objects~\cite{ding2025mosev2}. Following the semi-supervised VOS protocol, first-frame target masks are provided and the system predicts the same targets in the remaining frames. Table~\ref{tab:leaderboard} reproduces the official leaderboard results, with submission date and ID omitted for clarity; our team entry, \textbf{mmm}, is highlighted in bold.

\begin{table}[t]
\caption{Official MOSEv2 track leaderboard of the 8th LSVOS Challenge. Date and submission ID are omitted. The \textbf{mmm} row corresponds to our submission.}
\label{tab:leaderboard}
\centering
\begin{tabularx}{\textwidth}{@{}cl*{7}{>{\centering\arraybackslash}X}@{}}
\toprule
Rank & Participant &
$\mathcal{J}\&\mathcal{F}$ &
$\mathcal{J}$ &
$\dot{\mathcal{F}}$ &
$\mathcal{J}\&\mathcal{F}_{d}$ &
$\mathcal{J}\&\mathcal{F}_{r}$ &
$\mathcal{F}$ &
$\mathcal{J}\&\mathcal{F}$ \\
\midrule
1 & HITsz-Dragon & 69.82 & 68.21 & 71.43 & 79.12 & 34.87 & 73.76 & 70.99 \\
\textbf{2} & \textbf{mmm} & \textbf{66.20} & \textbf{64.79} & \textbf{67.60} & \textbf{81.57} & \textbf{26.88} & \textbf{69.74} & \textbf{67.26} \\
3 & kjeong & 64.37 & 63.16 & 65.59 & 80.26 & 27.53 & 67.09 & 65.12 \\
4 & venuszhou & 57.15 & 56.44 & 57.86 & 78.06 & 15.02 & 58.78 & 57.61 \\
5 & peterchao & 55.03 & 54.21 & 55.84 & 76.25 & 14.83 & 57.09 & 55.65 \\
6 & ygz & 54.87 & 54.18 & 55.57 & 72.15 & 14.50 & 56.76 & 55.47 \\
7 & ntulc & 53.27 & 52.52 & 54.03 & 71.83 & 11.90 & 55.85 & 54.18 \\
\bottomrule
\end{tabularx}
\end{table}

\subsection{Implementation Details}

We use SAM~3 as the base video segmentation model and fine-tune it for the MOSEv2 setting. Following our challenge configuration, training uses 8 NVIDIA A100 GPUs with batch size 1 and an input resolution of 1008. The learning rate is $3\times10^{-6}$ for the vision backbone and $5\times10^{-6}$ for the remaining modules. Competitive Memory Readout is applied during memory retrieval, and the adaptive restoration rule described in Sec.~\ref{sec:method} is used without an additional learned policy.

\subsection{Challenge Result and Qualitative Analysis}

As shown in Table~\ref{tab:leaderboard}, our method obtains a primary score of \textbf{66.20} and ranks \textbf{2nd}. Our submission also achieves 81.57 on the disappearance metric, while reappearance remains substantially more challenging. We therefore use the remaining report space mainly for qualitative visualisation of representative sequences.

\begin{figure}[t]
    \centering
    \includegraphics[width=\linewidth]{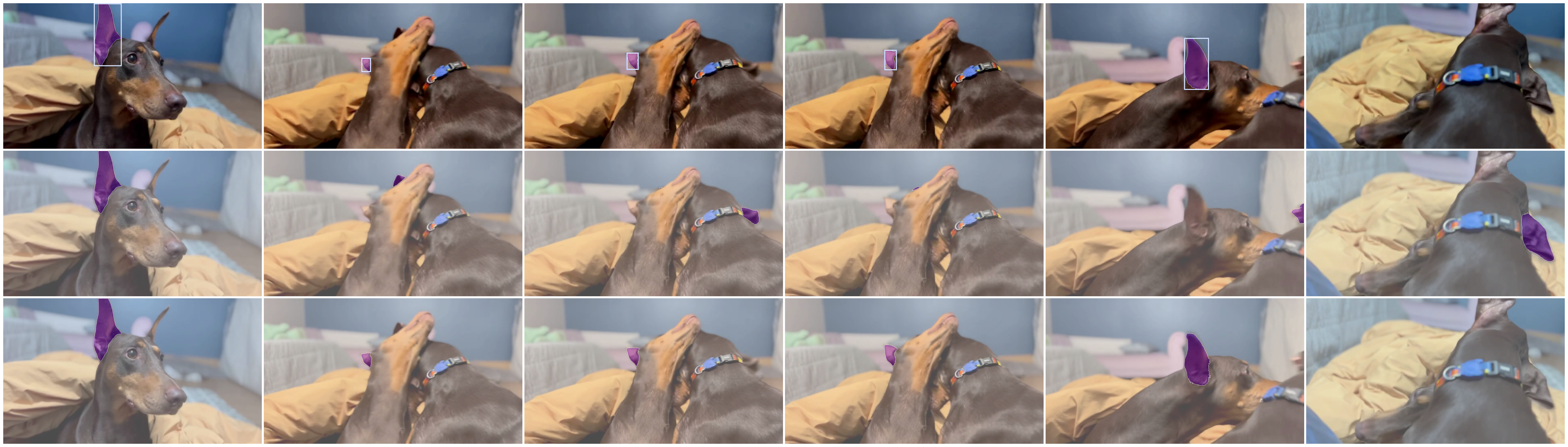}
    \caption{Qualitative comparison over a challenging long video sequence. 
    The rows show the ground truth, the SAM~3 baseline, and our method, respectively, with frames sampled from the beginning to the end of the sequence. 
    Compared with the baseline, our method maintains more stable target segmentation under long-term appearance variation and distractor interference.}
    \label{fig:qualitative}
\end{figure}

\section{Conclusion}
\label{sec:conclusion}

We presented our 2nd-place solution for the MOSEv2 track of the 8th LSVOS Challenge at ECCV 2026. The method augments SAM~3 with Competitive Memory Readout, explicitly comparing target memory against same-class competitor evidence to reduce identity drift. A simple adaptive restoration rule complements competition by recovering valid target evidence when suppression becomes too strong. The resulting method preserves the original SAM~3 tracking pipeline while improving robustness to the long-term occlusion, disappearance, reappearance, and similar-object interference that characterise MOSEv2. Our final submission achieves a primary challenge score of 66.20.

% ---- Bibliography ----
%
% BibTeX users should specify bibliography style 'splncs04'.
% References will then be sorted and formatted in the correct style.
%
\bibliographystyle{splncs04}
\bibliography{sam3_com_main}
\end{document}